\documentclass[nohyperref]{article}

\usepackage{microtype}
\usepackage{graphicx}
\usepackage{subfigure}
\usepackage{booktabs} 

\usepackage{hyperref}
\usepackage[accepted]{icml2022}

\PassOptionsToPackage{numbers, compress}{natbib}
\usepackage[utf8]{inputenc} 
\usepackage[T1]{fontenc}    
\usepackage{hyperref}       
\usepackage{url}            
\usepackage{booktabs}       
\usepackage{amsfonts}       
\usepackage{nicefrac}       
\usepackage{microtype}      
\usepackage{xcolor} 
\usepackage{soul}

\usepackage{times}
\usepackage{epsfig}
\usepackage{graphicx}
\usepackage{amsmath}
\usepackage{amsthm}
\usepackage{amssymb}
\usepackage[overload]{empheq}
\usepackage{bm}
\usepackage{float}
\usepackage{mathtools}

\newcommand{\eps}{\varepsilon}

\newcommand{\E}{\mathbb{E}}
\newcommand{\N}{\mathbb{N}}

\newcommand{\R}{\mathbb{R}}
\newcommand{\Acal}{\mathcal{A}}

\newcommand{\Ccal}{\mathcal{C}}

\newcommand{\Hcal}{\mathcal{H}}
\newcommand{\Kcal}{\mathcal{K}}

\newcommand{\Mcal}{\mathcal{M}}
\newcommand{\Pcal}{\mathcal{P}}

\newcommand{\Wcal}{\mathcal{W}}

\theoremstyle{definition}

\newcommand{\vM}{\mathbf{M}}
\newcommand{\vh}{\mathbf{h}}

\newcommand{\sN}{\mathbb{N}}

\newcommand{\What}{\widehat{\mathcal{W}}}

\begin{document}

\twocolumn[
\icmltitle{Conditional COT-GAN for Video Prediction with Kernel Smoothing}



\icmlsetsymbol{equal}{*}

\begin{icmlauthorlist}
\icmlauthor{Tianlin Xu}{xxx}
\icmlauthor{Beatrice Acciaio}{yyy}  

\end{icmlauthorlist}

\icmlaffiliation{xxx}{Department of Statistics, London School of Economics, London, UK}
\icmlaffiliation{yyy}{Department of Mathematics, ETH Zurich, Zurich, Switzerland}

\icmlcorrespondingauthor{Tianlin Xu}{t.xu12@lse.ac.uk}
\icmlcorrespondingauthor{Beatrice Acciaio}{beatrice.acciaio@math.ethz.ch}

\icmlkeywords{Machine Learning, ICML}

\vskip 0.3in
]
\printAffiliationsAndNotice{\icmlEqualContribution} 

\begin{abstract}
Causal Optimal Transport (COT) results from imposing a temporal causality constraint on classic optimal transport problems, which naturally generates a new concept of distances between distributions on path spaces.
The first application of the COT theory for sequential learning was given in \citet{DBLP:conf/nips/XuWMA20}, where COT-GAN was introduced as an adversarial algorithm to train implicit generative models optimized for producing sequential data. Relying on \cite{DBLP:conf/nips/XuWMA20}, the contribution of the present paper is twofold. First, we develop a conditional version of COT-GAN suitable for sequence prediction. This means that the dataset is now used in order to learn how a sequence will evolve given the observation of its past evolution. Second, we improve on the convergence results by working with modifications of the empirical measures via kernel smoothing due to \cite{pflug2016empirical}.  
The resulting \emph{kernel conditional COT-GAN} algorithm is illustrated with an application for video prediction. 
\end{abstract}

\section{Introduction}

Time series prediction is a challenging task. Given past observations, a desirable model should not only capture the distribution of features at each time step, but also predict its complex evolution over time. 
Autoregressive models which predict one time step after another seem to be a natural choice for learning such a task, see e.g. \cite{denton2018stochastic, kalchbrenner2017videopixel, oh2015action, weissenborn2019scaling}. However, the drawbacks of autoregressive models are the compounding error due to multi-step sampling and their high computational cost, see e.g. \cite{kalchbrenner2017videopixel, reed2017parallel}. 
Most existing models for time series prediction 
tend to ignore the temporal dependencies in sequences in the loss function, merely relying on certain specific network architectures, such as recurrent neural network (RNN) and 1D and 3D convolutional neural network (CNN), to capture the underlying dynamics, see e.g. \cite{srivastava2015unsupervised, aigner2018futuregan, tgan, vgan, mocogan}. 
For this learning task, the loss function used to compare prediction and real evolution plays a crucial role.
However, a loss function that is blind to the sequential nature of data will almost certainly disappoint.     

\citet{TimeGAN} proposed TimeGAN to tackle this problem by introducing  an auxiliary step-wise loss function to the original GAN objective, which indeed leads to more coherent and accurate predictions.
More recently, the advances in the field of causal optimal transport (COT) have shown a promising direction for sequential modeling, see e.g. \cite{backhoff2017causal, backhoff2020estimating, pflug2012distance, DBLP:conf/nips/XuWMA20}. This type of transport constrains the transport plans to respect temporal causality, in that the arrival sequence at any time $t$ depends on the starting sequence only up to time $t$. In this way, at every time we only use information available up to that time, which is a natural request in sequential learning.
This is the foundation of COT-GAN \cite{DBLP:conf/nips/XuWMA20}, where the training objective is tailored to sequential data. This proved to be an efficient tool, leading to generation of high-quality video sequences.  Although the sharpness of single frames remains a challenge in video modeling, COT-GAN demonstrates that the evolution of motions can be reproduced in a smooth manner without further regularization. 

While COT-GAN is trained to produce sequences, the algorithm we propose here is learning \emph{conditional sequences}, that is, how a sequence is likely to evolve given the observation of its past evolution.
For this task, we employ a modification of the empirical measure that was introduced by \citet{backhoff2020estimating} in the framework of \textit{adapted Wasserstein ($\Acal\Wcal$) distance}. $\Acal\Wcal$-distance is the result of an optimal transport problem where the plans are constrained to be causal in both direction (so-called \emph{bicausal optimal transport}); see \cite{pflug2012distance, pflug2016empirical}.
This turns out to be the appropriate distance to measure how much two processes differ, when we want to give importance to the evolution of information, see e.g. \cite{BBBE}.
As noted in \cite{pflug2016empirical} and \cite{backhoff2020estimating}, the $\Acal\Wcal$-distance between a distribution and the empirical measure of a sample from it may not vanish while the size of the sample goes to infinity. To correct for this, \citet{pflug2016empirical} proposed a convoluted empirical measure with a scaled smoothing kernel, while \citet{backhoff2020estimating} suggested an adapted empirical measure obtained by quantization - both aiming to smooth the empirical measure in some way in order to yield a better convergence.
In this paper, we follow the approach of adapting the empirical measure by kernel smoothing as done in \cite{pflug2016empirical}, and show that this smoothed empirical measure improves the performance of conditional COT-GAN. 

The process described above gives rise to \emph{kernel conditional COT-GAN}. The main contributions of the current paper can then be summarized as follows:

\begin{itemize}
    \item  we extend the COT-GAN to a conditional framework, powered by an encoder-decoder style generator structure; 
    \item we employ a new kernel empirical measure in the learning structure, which is a strongly consistent estimator with respect to COT;
    \item we show that our kernel conditional COT-GAN algorithm achieves state-of-the-art results for video prediction. 
\end{itemize}

\section{Framework}
We are given a dataset consisting of $n$ i.i.d. $d$-dimensional sequences  
$(x_1^i,\ldots,x^i_T)_{i=1}^n$ where $T\in\sN$ is the number of time steps and $d\in\sN$ is the dimensionality at each time.  This is thought of as a random sample from an underlying distribution $\mu$ on $\mathbb{R}^{d\times T}$, from which we want to extract other sequences. More precisely, we want to learn the conditional distribution of $(x_{k+1},\ldots,x_T)$ given $(x_1,\ldots,x_k)$ under $\mu$, for any fixed $k\in\{1,\ldots,T-1 \}$. In the application of video prediction, an entire video contains $T$ frames, each of which has resolution $d$. The first $k$ frames of the video are taken as an input sequence, and later frames from time $k+1$ to $T$ are the target sequence. 
We will use the notation $x_{s:t} = (x_s, ..., x_t)$, for $1 \leq s\leq t \leq T$.

The conditional learning will be done via a conditional generative adversarial structure, based on a specific type of optimal transport tailored for distributions on path spaces, as introduced in the next section, in the wake of what is done in \cite{DBLP:conf/nips/XuWMA20}. 


\section{Optimal Transport and Causal Optimal Transport} \label{section:COT}
Given two probability measures 
$\mu,\nu$ defined on $\R^D$, $D\in\N$, and a cost function $c:\R^D\times\R^D\to\R$, 
the classical (Kantorovich) optimal transport of $\mu$ into $\nu$ is formulated as
\begin{equation}\label{Wc_primal}
\Wcal_c(\mu,\nu):=
\inf_{\pi\in\Pi(\mu,\nu)}\E^{\pi}[c(x,y)],
\end{equation}
where $\Pi(\mu,\nu)$ is the set of probability measures on $\R^D\times\R^D$ with marginals $\mu,\nu$, which are called transport plans between $\mu$ and $\nu$. Here $c(x,y)$ is interpreted as the cost of transporting a unit of mass from $x$ to $y$. $\Wcal_c(\mu,\nu)$ is thus the minimal total cost to transport the mass $\mu$ to $\nu$. When $c(x,y)$ is a distance function between $x$ and $y$ (usually $\|x-y\|_p$ for some $1\leq p<\infty$), $\Wcal_c(\mu,\nu)$ is known as Wasserstein distance or Earth mover distance.

We are interested in transports between path spaces, that is, $D=d\times T$ in the above notations. Since now there is a time component intrinsic in the space $\R^D$, we are adopting a particular kind of transport which is tailored for path spaces.  
We denote by $x=(x_1,...,x_T)$ and $y=(y_1,...,y_T)$ the first and second half of the coordinates on $\R^{d\times T}\times\R^{d\times T}$, respectively. A probability measure $\pi$ on $\R^{d\times T}\times\R^{d\times T}$ is called \emph{causal transport plan} if it satisfies the constraint
\begin{equation} \label{eq:causal_pi}
\pi(dy_t|dx_{1:T})=\pi(dy_t|dx_{1:t})\qquad \text{for all\, $t=1,\cdots,T-1$}.
\end{equation}
Intuitively, the probability mass moved to the arrival sequence at time $t$ only depends on the starting sequence up to time $t$.
The set of causal plans between $\mu$ and $\nu$ is denoted by $\Pi^{\Kcal}(\mu,\nu)$, and
restricting the space of transport plans in \eqref{Wc_primal} to such a set gives rise to the causal optimal transport problem:
\begin{equation}\label{eq:COT}
\Wcal^\Kcal_c(\mu, \nu) := \inf_{\pi \in \Pi^{\Kcal}(\mu, \nu)} \E^{\pi}[c(x,y)].
\end{equation}
COT has already found wide application in dynamic problems in stochastic calculus and mathematical finance, see e.g. \cite{ABZ,ABC,ABJ,BBBE,backhoff2020estimating}, and first numerical results are given in \cite{ABJ,DBLP:conf/nips/XuWMA20}.

\section{COT-GAN and CCOT-GAN} \label{section:cCOT-GAN}
In this section we will recall the main steps that led to the COT-GAN algorithm for sequential learning in \citet{DBLP:conf/nips/XuWMA20}, and refer to Appendix A for the details. We then introduce a conditional version, called \emph{conditional} COT-GAN (CCOT-GAN), suited for sequential prediction.

Solving (causal) optimal transport problems is typically computational costly for large datasets. One way to circumvent this challenge is to resort to approximations of transport problems by means of efficiently solvable auxiliary problems. Notably, \citet{GPC} proposed the \emph{Sinkhorn divergence}, which allows for the use of the Sinkhorn algorithm \cite{cuturi2013sinkhorn}. The first observation is that \eqref{Wc_primal} is the limit for $\eps\to 0$ of the entropy-regularized transport problems
\begin{equation}\label{eq:Wce}
\Pcal_{c,\eps}(\mu,\nu) := \inf_{\pi\in\Pi(\mu,\nu)}\{\E^\pi[c(x,y)]-\eps H(\pi)\},\quad \eps>0,
\end{equation} 
where $H(\pi)$ is the Shannon entropy of $\pi$. 
Denoting by $\pi_{c,\eps}(\mu,\nu)$ the optimizer in \eqref{eq:Wce}, and by $\mathcal{W}_{c,\eps}(\mu,\nu):=\E^{\pi_{c,\eps}(\mu,\nu)}[c(x,y)]$  the resulting total cost, the Sinkhorn divergence is defined as
\begin{equation}\label{Sink}
\widehat{\mathcal{W}}_{c,\eps}(\mu,\nu):= 
2\mathcal{W}_{c,\eps}(\mu,\nu)-\mathcal{W}_{c,\eps}(\mu,\mu)-\mathcal{W}_{c,\eps}(\nu,\nu).
\end{equation}

Similarly, in a causal setting, we consider the entropy-regularized COT problems
\begin{equation}\label{eq:Wcek}
\Pcal^\Kcal_{c,\eps}(\mu,\nu) := \inf_{\pi\in\Pi^\Kcal(\mu,\nu)}\{\E^\pi[c(x,y)]-\eps H(\pi)\} ,\quad \eps>0, 
\end{equation}
approximating \eqref{eq:COT}. 
By using an equivalent characterization of causality (see Appendix A),  this
can be reformulated as a maximization over regularized transport problems with respect to a specific family of cost functions:
\begin{equation}\label{Pcce}
\Pcal^\Kcal_{c,\eps}(\mu,\nu) = \sup_{c^\Kcal\in\Ccal^\Kcal(\mu,c)}  \Pcal_{c^\Kcal,\eps}(\mu,\nu).
\end{equation}
The family of costs $\Ccal^\Kcal(\mu,c)$ is given by
\begin{align}\label{eq:L_set}
\nonumber \Ccal^\Kcal(\mu,c):= & \Bigg\{c(x,y) + \sum_{j=1}^J \sum_{t=1}^{T-1} h^j_t(y)\Delta_{t+1}M^j(x): \\
& J\in\N, (h^j,M^j)\in\Hcal(\mu) \Bigg\},
\end{align}
where $\Delta_{t+1}M(x) := M_{t+1}(x_{1:t+1}) - M_t(x_{1:t})$ and $\Hcal(\mu)$ is a set of functions depicting causality:
\begin{align*}
\Hcal(\mu):= & \{(h,M) : h=(h_t)_{t=1}^{T-1}, h_t\in\Ccal_b(\R^{d\times t}),\\           
& M=(M_t)_{t=1}^{T}\in\Mcal(\mu),\  M_t\in \Ccal_b(\R^{d\times t})\},
\end{align*}
with
$\Mcal(\mu)$ being the set of martingales on $\R^{d\times T}$ w.r.t. the canonical filtration and the measure $\mu$, and  $\Ccal_b(\R^{d\times t})$ the space of continuous, bounded functions on $\R^{d\times t}$.
This suggests the following as a robust version of the Sinkhorn divergence from (\ref{Sink}) that takes into account causality:
\[
\sup_{c^\Kcal\in\Ccal^\Kcal(\mu,c)}\widehat{\mathcal{W}}_{c^\Kcal,\eps}(\mu,\nu).
\]
This is the distance used by the discriminator in COT-GAN \cite{DBLP:conf/nips/XuWMA20}  
in order to evaluate the discrepancy between real data and generated one (up to a slightly different definition of Sinkhorn divergence, see Appendix A), and it is the one we will use in the current paper for sequential prediction.

Furthermore, \cite{DBLP:conf/nips/XuWMA20}
makes the two following adjustments
needed to make computations feasible. First, rather than considering the whole set of costs in \eqref{eq:L_set}, in \eqref{Pcce} we optimize over a subset $\Ccal^\Kcal(\mu,c)$, by considering $\vh:=(h^j)_{j=1}^J$ and $\vM:=(M^j)_{j=1}^J$ of dimension bounded by a fixed $J\in\mathbb{N}$.
Second, instead of requiring $\vM$ to be a martingale, we consider all continuous bounded functions and introduce a regularization term which penalizes deviations from being a martingale.
For a mini-batch of size $m$, $\{x^i_{1:T}\}_{i=1}^m$, sampled from the dataset, 
the martingale penalization for $\vM$ is defined as
\[
{p}_{\vM}(\widehat{\mu}):=\frac{1}{mT}\sum_{j=1}^J\sum_{t=1}^{T-1}\Bigg|\sum_{i=1}^m \frac{M^j_{t+1}(x^i_{1:t+1}) - M^j_t(x^i_{1:t})}{\sqrt{\text{Var}[M^j]} + \eta}\Bigg|,
\]
where $\widehat{\mu}$ is the empirical measure corresponding to the mini-batch sampled from the dataset, $\text{Var}[M]$ is the empirical variance of $M$ over time and batch, and $\eta>0$ is a small constant.
This leads to the following objective function for COT-GAN in \cite{DBLP:conf/nips/XuWMA20}:
\begin{equation}\label{eq:minibath_objective}
\What_{c^\Kcal_{\varphi}, \eps}(\widehat{\mu},\widehat{\nu}_\theta) - {\lambda} p_{{\bf M}_{\varphi_2}} (\widehat{\mu}),
\end{equation}
where $\widehat{\nu}_\theta$ is the empirical measure corresponding to the mini-batch produced by the generator, parameterized by $\theta$, $\vh_{\varphi_1}$ and $\vM_{\varphi_2}$ represent the discriminator who learns the worst-case cost $c^\Kcal_{\varphi}$, parameterized by  $\varphi:=(\varphi_1, \varphi_2)$, and $\lambda$ is a positive constant (see Appendix A for details).

We now extend the analysis developed in \cite{DBLP:conf/nips/XuWMA20} to a conditional framework for sequence prediction.
Given the past history of a sequence up to time step $k$, the aim of CCOT-GAN is learning to predict the evolution from time step $k+1$ to $T$.
The learning is done by stochastic gradient descent (SGD) on mini-batches. 
Given a sample $\{{x}^i_{1:T}\}_{i=1}^{m}$ from the dataset and a sample
$\{ {z}^i_{k+1:T} \}^m$ from a distribution $\zeta$  (noise) on some latent space $\mathcal{Z}$,   we define the generator as a conditional model $g_\theta$, parameterized by $\theta$, which predicts
the future evolution
$\hat{x}_{k+1:T}^i = g_{\theta}({x}^i_{1:k}, {z}^i_{k+1:T})$. 
The prediction $\hat{x}_{k+1:T}^i$ is then concatenated with the corresponding input sequence ${x}^i_{1:k}$ over the time dimension in order to be compared with the training sequence ${x}^i_{1:T}$  by the discriminator.
We denote the empirical distributions of real and concatenated data by
\[
\widehat{\mu}:=\frac{1}{m}\sum_{i=1}^m\delta_{{x}^i_{1:T}},\quad
\widehat{\nu}^c_\theta:=\frac{1}{m}\sum_{i=1}^m\delta_{\text{concat}({x}^i_{1:k},\hat{x}_{k+1:T}^i)},
\]
where $\widehat{\nu}^c_\theta$ incorporates the parameterization of $g_\theta$ through $\{\hat{x}_{k+1:T}^i\}_{i=1}^m$.
Following COT-GAN's formulation of adversarial training, we arrive at the parameterized objective function for CCOT-GAN: 
\begin{equation} \label{eq:ccotgan_obj}
\What_{c_{\varphi}^\Kcal, \eps}(\widehat{\mu}, \widehat{\nu}^c_\theta) - {\lambda} p_{{\bf M}_{\varphi_2}} (\widehat{\mu}).
\end{equation}
In the implementation of CCOT-GAN, the generator $g_\theta$ is broken down into two components: an encoder that learns the features of input sequences $\{ {x}^i_{1:k} \}_{i=1}^{m}$ and a decoder that predicts future evolutions given the features of inputs and noise $\{ {z}^i_{k+1:T} \}_{i=1}^{m}$. 
The discriminator role is played by $\vh_{\varphi_1}$ and $\vM_{\varphi_2}$, which are parameterized separately by two neural networks that respect temporal causality.
These can take the shape of RNNs or 1D or 3D CNNs that are constrained to causal connections only, see Appendix B for details.
We maximize the objective function \eqref{eq:ccotgan_obj} over $\varphi$ to search for a robust (worst-case) distance between the two empirical measures $\widehat{\mu}$ and $\widehat{\nu}^c_\theta$, and minimize it over $\theta$ to learn a conditional model that produces sequential prediction. 

\section{Adapted Empirical Measure and KCCOT-GAN}\label{sect.aem}
It was noted by \citet{backhoff2020estimating} and \citet{pflug2016empirical} that
the (classical) empirical measures are not necessarily consistent estimators with respect to distances originating from transport problems where transports plans respect causality constraints. The \emph{nested distance} \cite{pflug2012distance} or \emph{adapted Wasserstein} distance \cite{backhoff2020estimating} is the result of an optimal transport problem where plans are required to satisfy the causality constraint \eqref{eq:causal_pi} as well as its symmetric counterpart, when inverting the role of $x$ and $y$:
\begin{align}
\nonumber \Acal\Wcal_c(\mu,\nu):= \inf \{ \E^{\pi}[c(x,y)] : & \pi\in\Pi^\Kcal(\mu,\nu), \\ 
& \pi'\in\Pi^\Kcal(\nu,\mu) \}, \label{AW}
\end{align}
where $\pi'(dx,dy)=\pi(dy,dx)$. 

Now, for any measure $\mu$, and for the empirical measures  $\widehat\mu_m$ relative to a random sample of size $m$ from it, it is known (see e.g. \cite{fournier2015rate}) that
\[
\Wcal_c (\mu, \widehat\mu_m) \rightarrow 0\quad \text{as $m\to\infty$},
\]
whereas
\cite{backhoff2020estimating, pflug2016empirical} observe that this is not necessarily true when substituting the Wasserstein distance $\Wcal_c$ with the adapted Wasserstein distance $\Acal\Wcal_c$.
This is of course undesirable, in particular thinking of the fact that the discriminator will evaluate discrepancies  between real and generated measures by relying on empirical measures of 
the corresponding minibatches, see Section \ref{section:cCOT-GAN} and \cite{DBLP:conf/nips/XuWMA20}.

In \cite{backhoff2020estimating} and \cite{pflug2016empirical}, two different ways of adapting the empirical measure are suggested: by smoothing using a scaled kernel and by a quantization technique, respectively.
The quantization technique\cite{backhoff2020estimating} divides the data space into sub-cubes, and maps every value to the center of the sub-cube to which it belongs. We did not adopt this approach for two reasons: first, the convergence property proved in Theorem 1.3 in \cite{backhoff2020estimating} only holds when the number of sub-cubes is extremely small if the dimensionality of the data is large (typically a few hundreds). To see why too few sub-cubes can be problematic, consider this technique with two sub-cubes. This will map all data into only two possible values, which discards substantial information from the original data.  Second, the quantization technique is non-differentiable, requiring an approximation so the gradients can flow back via back-propagation in the stage of learning. We therefore adopt the kernel smoothing approach which
we describe in detail in the remainder of this section.  

For a probability measure $\mu$ with density $f$, and a density function $k_h (x) := \frac{1}{h} k(\frac{x}{h})$ where $h$ is the bandwidth parameter, the density estimator $\hat{f}$ is defined as 
\begin{align}
 \hat{f} (x)  = \int k_h(x-y) f(y) dy = f * k_h(x),
\end{align}
where $ * $ denotes the convolution of densities. 

Denoting the measure induced by density $k_h$ as $K^f$, we can write the convoluted 
measures with density $k_h$ as the weighted empirical measures of $\widehat{\mu}$ and $\widehat{\nu}^c_\theta$:
\begin{align}
\widehat{\mu}^f & := \widehat{\mu}* K^f = \sum_{i=1}^m w_i \delta_{{x}^i_{1:T}}, \\
\widehat{\nu}^{c, f}_\theta & := \widehat{\nu}^c_\theta * K^f = \sum_{i=1}^m w_i \delta_{\text{concat}({x}^i_{1:k},\hat{x}_{k+1:T}^i)},
\end{align}
where the weight $w_i$ is determined by $k_h$. Intuitively, this smooths the observations by taking a weighted average of all observations, typically with more influence from neighboring points.

\citet{pflug2016empirical} proved that the adapted Wasserstein distance of the convoluted measures converges, i.e., 
\begin{equation*}
    P(\Acal\Wcal_c(\widehat{\mu}^f, \widehat{\nu}^{c, f}_\theta) > \eps) \rightarrow 0 \quad \text{as  } m \rightarrow \infty,
\end{equation*}
provided that 
\textit{
\begin{enumerate}
  \item the kernel $k_h$ is nonnegative and compactly supported on $\mathbb{R}^D$,
  \item the density f is bounded and uniformly continuous,
  \item the bandwidth $h$  is a function of the sample size $m$ that satisfies 
  \begin{align}\label{eq:h_cons}
      \nonumber h_m \rightarrow 0, & \quad \frac{mh_m}{|\log h_m|} \rightarrow \infty, \quad \frac{|\log h_m|}{\log \log m} \rightarrow \infty, \\ 
      & \textit{and} \quad mh_m \rightarrow \infty, \quad \text{as} \quad  m \to \infty,
  \end{align}
  \item the measures $\mu$ and $\nu$ are conditionally Lipschitz.
\end{enumerate}
}
For proofs and detailed discussions, please see Theorem 2 and 4 in \cite{pflug2016empirical}.

Note that convergence result above is derived for the adapted Wasserstein distance $\Acal\Wcal_c$. In order to deduce the results on $\Wcal^\Kcal_c$, notice that
\begin{equation}\label{eq.dis}
\Wcal^\Kcal_c(\mu,\nu)\leq  \Acal\Wcal_c(\mu,\nu)
\end{equation}
for any probability measures $\mu,\nu$ and any cost function $c$, given that the set of transports over which minimization is done for causal optimal transport is bigger than that for $\Acal\Wcal$-distance, cf. \eqref{eq:COT} and \eqref{AW}.

Relying on this convergence result, we now introduce the CCOT-GAN with kernel smoothing (KCCOT-GAN).
The objective function of KCCOT-GAN at the level of minibatches is computed on the
adapted empirical measures:
\begin{equation} \label{eq:qccotgan_obj}
\What_{c_{\varphi}^\Kcal, \eps}(\widehat\mu^f, \widehat\nu^{c,f}_\theta) - {\lambda} p_{{\bf M}_{\varphi_2}}(\widehat\mu^f).
\end{equation}
We maximize the objective function over $\varphi$ to search for a worst-case distance between the two adapted empirical measures, and minimize it over $\theta$ to learn a conditional distribution that is as close as possible to the real distribution. The algorithm is summarized in Algorithm \ref{alg}. Its time complexity scales as $\mathcal{O}((J+2d)2LTm^2)$ in each iteration. The distance $\What_{c_{\varphi}^\Kcal, \eps}(\widehat\mu^f, \widehat\nu^{c,f}_\theta)$ is approximated by the means of the Sinkhorn algorithm iteratively with a fixed number of iterations, see 
Appendix A. 

\begin{algorithm}[tb]
   \caption{training KCCOT-GAN by SGD}
   \label{alg}
\begin{algorithmic}
   \STATE {\bfseries Input:} $\{{x}^i_{1:T}\}_{i=1}^{n}$(data), $\zeta$(distribution on latent space)
   \STATE {\bfseries Parameters:} $\theta_0$, $\varphi_{0}$(initialization of parameters), $m$(batch size), $\eps$(regularization parameter),  
   $\alpha$(learning rate), $\lambda$(martingale penalty coefficient), $h$(bandwidth parameter)
   \REPEAT
   \STATE (1) Sample $\{{x}^i_{1:T}\}_{i=1}^{m}$ from real data;
   \STATE (2) Learn features from input sequences: \\
   $\qquad \quad \{{e}^i_{1:T}\}_{i=1}^{m} \leftarrow f_{\theta_e}(\{{x}^i_{1:T}\}_{i=1}^{m})$;
   \STATE (3) Sample $\{{z}^i_{k:T-1}\}_{i=1}^{m}$ from $\zeta$;\\ 
   \STATE (4) Predict conditioned on features and inputs: \\
   $\{\hat{x}_{k+1:T}^i\}_{i=1}^{m}$ \\ 
   $\qquad \quad \leftarrow f_{\theta_d}(\{{e}^i_{1:T}\}_{i=1}^{m}, \{{x}^i_{k:T-1}\}_{i=1}^{m}, \{{z}^i_{k:T-1}\}_{i=1}^{m})$;
   \STATE (5) Obtain smoothed measures: $\widehat{\mu}^f$ and $\widehat{\nu}^{c, f}_\theta $;
   \STATE (6) Compute $\What_{c_{\varphi}^\Kcal, \eps} (\widehat{\mu}^{f}, \widehat{\nu}_\theta^{c,f})$ by the Sinkhorn algorithm;
   \STATE (7) Update discriminator parameter: \\
   ${\varphi} \leftarrow {\varphi} + \alpha \nabla_\varphi\Big( \What_{c_{\varphi}^\Kcal, \eps}(\widehat{\mu}^f, \widehat{\nu}_\theta^{c,f}) - {\lambda} p_{{\bf M}_{\varphi_2}}(\widehat\mu^f)\Big)$; \\
   \STATE (8) Repeat step (2) - (6);
  \STATE (9) Update generator parameter: \\
  $\qquad \theta \leftarrow \theta - \alpha \nabla_\theta\left( 
    \What_{c_{\varphi}^\Kcal, \eps}(\widehat{\mu}^f, \widehat{\nu}_\theta^{c,f})\right)$;
  \UNTIL{convergence}
\end{algorithmic}
\end{algorithm}

\section{Implementation of KCCOT-GAN}
The generator of KCCOT-GAN consists of an encoder that learns features from the input sequences, and a decoder that generates predictions conditioned on the input features and noise, supported by convolutional LSTM (convLSTM)\cite{shi2015convlstm}. 
The decoder was trained using a hierarchical version of the Teacher Forcing algorithm \cite{williams1989learning} which feeds the real values from observations as inputs during the training stage, in order to reduce the compounding error from multi-step predictions. To make it concrete, we proceed to formulate the implementation of KCCOT-GAN. 

To avoid confusion, we refer to the entire input $x_{1:T}$ as the input sequence, and to the sequence $x_{1:k}$ upon which the prediction $x_{k+1:T}$  is made as the context sequence.
Since the full input sequence is available to us at the stage of training,  we first learn the hierarchical features of it through an encoder with $n$ layers,
\begin{align*}
    e^1_{1:T} & = f_{\theta^1_e}(x_{1:T}), \\
    e^2_{1:T} & = f_{\theta^2_e}(e^1_{1:T}), \\
    & \quad \vdots \\
    e^n_{1:T} & = f_{\theta^n_e}(e^{n-1}_{1:T}).
\end{align*}
From here on, we denote the encoder as $f_{\theta_e}$ parametrized by $\theta_e := \{\theta^1_e, \theta^2_e,..., \theta^n_e\}$, and the features extracted by the encoder as $e_{1:T} := \{ e_{1:T}^1, ..., e_{1:T}^n\}$. 

To deploy the teacher forcing algorithm, we make use of the hierarchical features as well as the input sequence. 
At time step $k+1$, we predict $\hat{x}_{k+1}$ conditioned on $(e_{k}, x_{k})$, under the assumption that the feature $e_k$ contains all the information about the context sequence.  Instead of feeding the prediction $\hat{x}_{k+1}$ back to the model to make next prediction, we continue to predict $\hat{x}_{k+2}$ conditioned on $(e_{k+1}, x_{k+1})$ in an effort to prevent the model to derail from the truth by making a mistake in an intermediate step.  
As a result, we train the model to predict $\hat{x}_{k+1:T}$ conditioned on ($e_{k:T-1}$, $x_{k:T-1}$).  In the inference stage, however, we do not have the information beyond the context sequence. The prediction is therefore completed in an auto-regressive manner.

Given Gaussian noise $z_{k:T-1}$, the decoder $f_{\theta_d}$ with $l$ layers for $l \geq n+1$ learns to predict the future steps by 
\begin{align*}
    d^1_{k+1:T} & = f_{\theta^1_d}(e^n_{k:T-1}, z_{k:T-1}), \\
    & \quad \vdots \\
    d^{l-1}_{k+1:T} & = f_{\theta^{l-1}_d}(e^{1}_{k:T-1}, d^{l-2}_{k+1:T}) \\
    \hat{x}_{k+1:T} & = f_{\theta^l_d}(x_{k:T-1}, d^{l-1}_{k+1:T}).
\end{align*}

As usual, the generator parameters $\theta := \{\theta_e, \theta_d\}$ and discriminator parameters $\varphi$ are learned on the level of mini-batches via Stochastic Gradient Descent (SGD).  To yield better convergence property, we smooth the mini-batches in each iteration using a scaled Gaussian kernel with zero mean, 
\begin{equation*}
    k_h(x) = \frac{1}{h} e^{-\frac{x^2}{2h^2}}.
\end{equation*}

Differently from the technique of Gaussian blur widely used in image processing, see e.g. \cite{haddad1991class, reinhard2010high, nixon2019feature, getreuer2013survey}, we apply a 3D scaled Gaussian kernel to both spatio and temporal dimensions. In another line of work, \citet{zhang2020spread} show that convoluting measures with a kernel density estimator is also a valid approach to tackle the problem of disjoint supports in divergence minimization. 

The choices of the bandwidth parameter $h$ are restricted by the conditions in Eq. \eqref{eq:h_cons}. In the implementation, we relax this assumption by deploying a decaying bandwidth as a function of the number of the training iterations, rather than a function of sample size $m$. We realize that this simplification may lead to inferior theoretical guarantee of convergence. However, we will leave the exploration of a more appropriate approach to satisfy the theoretical assumptions to future research.

\section{Related Work}
Video prediction is an active area of research.  Methods relying on Variational inference\cite{blei2017variational} and VAE \cite{kingma2013vae}, e.g. SV2P \cite{babaeizadeh2017stochastic}, SVP-LP  \cite{denton2018stochastic}, VTA  \cite{kim2019variational}, and VRNN \cite{castrejon2019improved}, have shown promising results.  
The majority of adversarial models adopted in this domain were trained on the original GAN objective \cite{goodfellow2014generative} or the Wasserstein GAN objective \cite{arjovsky2017wasserstein}, both of which provide step-wise comparison of sequences. SAVP \cite{lee2018stochastic} combined the objective function of the original GAN and VAE to achieve the state of the art performance.  

Substantial efforts have been devoted to designing specific architectures that tackle the spatio-temporal dependencies, e.g. \citep{vgan, tgan, mocogan, clark2019adversarial, mathieu2015deep, villegas2017decomposing}, and training schemes that facilitate learning, e.g. \cite{mathieu2015deep, villegas2017decomposing, aigner2018futuregan}. Whilst some works such as TGAN \cite{tgan} and VGAN \citep{vgan} combined a static content generator with a motion generator, others, e.g. \citep{mocogan, clark2019adversarial}, designed two discriminators to evaluate the spatial and temporal components separately.   
\citet{mathieu2015deep} explored a loss that measures gradient difference at frame level on top of an adversarial loss trained with a multi-scale architecture. As a result, better performance was achieved in comparison to a simple mean square error loss commonly used in the literature. MCnet  \cite{villegas2017decomposing} extended  \cite{mathieu2015deep} by adopting \emph{convolutional long short-term memory (ConvLSTM)} \cite{shi2015convlstm} in the networks. Alternatively, 3D CNN with progressively growing training scheme \cite{karras2017pggan} was also shown to be successful by FutureGAN  \cite{aigner2018futuregan}.

However, it may not be sufficient to rely solely on the network architecture to capture the temporal structure of data.
An important development in time series synthesis and prediction is the identification of more suitable loss functions. TimeGAN \cite{TimeGAN} combined the original GAN loss with a step-wise loss that computes the distance between the conditional distributions in a supervised manner. By matching a conditional model to the real conditional probability $p(x_t|x_{1:t-1})$ at every time step, it explicitly encouraged the model to consider the temporal dependencies in the sequence. 
In comparison, COT-GAN \cite{DBLP:conf/nips/XuWMA20} explored a more natural formulation for sequential generation which leads to convincing results.  

\section{Experiments} \label{sec:experiments}
\begin{figure*}[t!]
    \centering
    \includegraphics[width=1.0\textwidth]{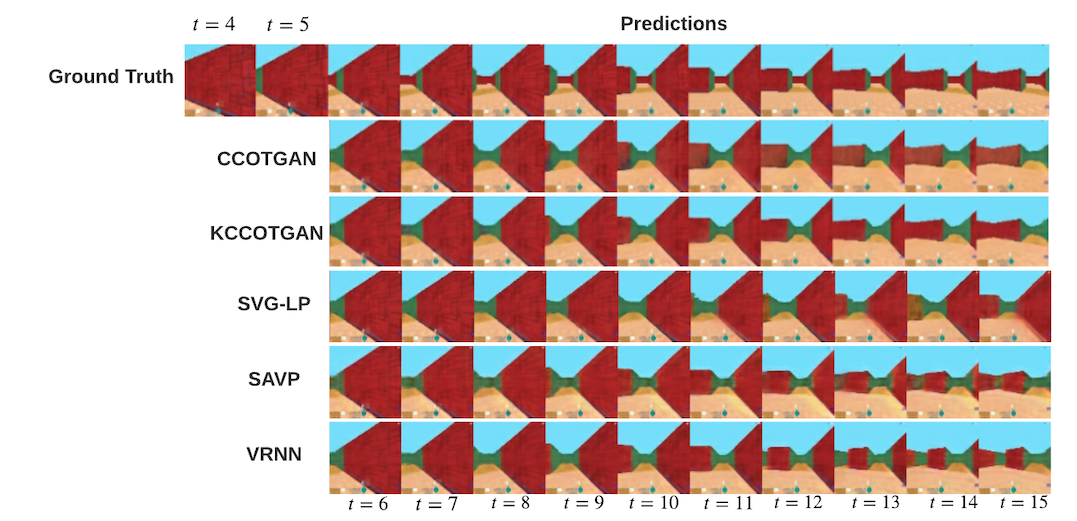}
    \caption{GQN Mazes results on the test set. Only the last 2 frames from the context sequence are shown. }
    \label{fig:mazes}
\end{figure*} 
We compare \textbf{KCCOT-GAN} to \textbf{CCOT-GAN} without kernel smoothing as an ablation study, to \textbf{SVP-LP} (\citet{denton2018stochastic}), to \textbf{SAVP} (\citet{lee2018stochastic}), and to \textbf{VRNN} (\citet{castrejon2019improved}), on three well-established video prediction datasets.  The source code and video results are available at \url{https://github.com/neuripss2020/kccotgan}. 
In all our experiments, the choice of cost function is $c(x,y)=\sum_t\|x_t-y_t\|_2^2$, and initial bandwidth $h$ is 1.5 and is gradually decayed to 0.1 as training progresses.  We select the first 15 frames and downsample them to a resolution of $64 \times 64$. We use the first 5 frames as the context sequence and the rest 10 frames as the target sequence. All results are evaluated on test sets. Note that the maximum number of hidden units used for the layers in the generator and discriminator networks is 256 for the \textbf{GQN Mazes} and \textbf{BAIR Push Small} datasets and 128 for the \textbf{Moving MNIST} dataset, due to the constraint of available computation power. This is at most half of the baseline model sizes.  Although a compromised model capacity is expected, KCCOT-GAN still produces excellent results on various tasks.  
Network architectures and more training details are given in Appendix B.

\paragraph{GQN Mazes.}
The GQN Mazes was first introduced by \cite{eslami2018neural} for training agents to learn their surroundings by moving around.  The dataset contains random mazes generated by a game engine.  A camera traverses one or two rooms with multiple connecting corridors in each maze. 
The dataset comes with a training set that contains 900 sequences and a test set with a size of 120.   The original sequences have a length of 300 and resolution of 84 $\times$ 84.

Figure \ref{fig:mazes} demonstrates that all models successfully captured the spatial structure in the frames well. However, predictions produced by SVG-LP lack of the evolution of motions, which is observed in many reproduced results of the model across various dataset. This could be attributed to the fact that SVG-LP is conditioned on a single frame from the previous time step, which makes it impossible for the model to pick up any information about past evolution. Visually, KCCOT-GAN and VRNN produced the sharpest frames out of all. Whilst samples from VRNN show more variations, those from KCCOT-GAN tend to be closer to the ground truth which may contribute to the better numerical evaluations in Table \ref{tab:compare1}. 

\paragraph{BAIR Push Small.}
Due to computation and storage constraint, we opted for this smaller version of the original BAIR Push dataset. The BAIR Push Small contains about 44,000 example with a resolution of $64 \times 64$. Each example shows a sequence of motions of robot arm pushing objects on a table.

\begin{figure*}[t!]
    \centering
    \includegraphics[width=1.0\textwidth]{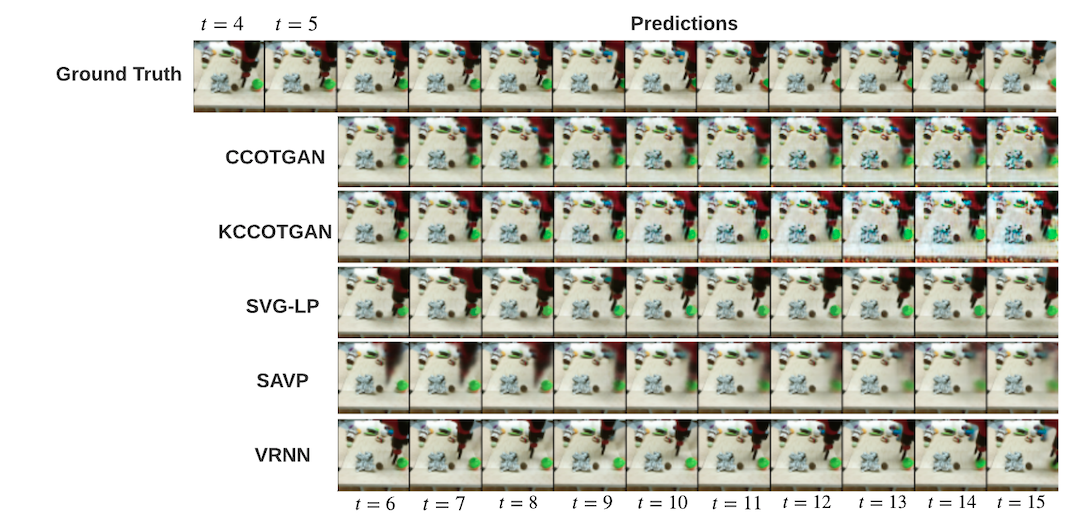}
    \caption{BAIR Push Small results on the test set. Only the last 2 frames from the context sequence are shown.}
    \label{fig:bairpush}
\end{figure*}

For this dataset, the results from SVG-LP and VRNN are extremely good in terms of both the image quality and the variation in samples, see Figure \ref{fig:bairpush}. It is clearly a very difficult task to outperform these two baselines.  On the other hand, SAVP has failed in producing high quality predictions.

On this dataset, although KCCOT-GAN underperforms the SVG-LP and VRNN baselines, we observe a clear improvement in sharpness from CCOT-GAN to KCCOT-GAN.
As these two models share the same network structure and hyper-parameter settings, we can confirm that this improvement solely comes from the adaption of empirical measures via kernel smoothing. 

\paragraph{Moving MNIST Dataset.}
Moving MINST\cite{srivastava2015unsupervised} contains two digits that move with
velocities sampled uniformly in the range of 2 to 6 pixels
per frame and bounce within the edges of each frame. The dataset has 10000 sequences, of which we use 8000 for training and the rest for testing. Each of the original sequence contains 20 frames with resolution $64 \times 64$. Results are given in Table \ref{tab:compare1} and Appendix C. 

\paragraph{Evaluation.} 
We evaluate the video predictions using three metrics: 
Structural Similarity index\cite{wang2004ssim} (SSIM, higher is
better),
Learned Perceptual Image Patch Similarity\cite{zhang2018lpips} (LPIPS, lower
is better), 
Fr\'echet Video Distance\cite{fvd} (FVD, lower is better).

The evaluation scores are reported in Table \ref{tab:compare1}.  We can see that KCCOT-GAN outperforms the baseline models on GQN Mazes dataset based on the three metrics.  However, VRNN are well ahead other models in BAIR Push Small dataset.  The performances of VRNN and KCCOT-GAN on the Moving MMNIST dataset is reasonably close with KCCOT-GAN leading in SSIM and LPIPS but VRNN having better FVD score.
\begin{table}[!h]
\centering
\caption{Evaluations for video datasets. Lower values in the metrics indicate better sample quality for LPIPS and FVD, whereas higher values in SSIM are better.}
\begin{tabular}[!t]{lcccc}
\hline
\hline
\bf{GQN Mazes}      &  SSIM & LPIPS  &  FVD  \\
\hline
SAVP     &  0.49   & 0.077   &  488.35    \\
VRNN   &  0.56   &  0.062  &   345.51   \\
SVG-LP   &  0.43   & 0.094    & 575.22  \\
CCOT-GAN     &  0.60   & 0.061   &  323.28     \\
KCCOT-GAN &  \bf{0.64}   & \bf{0.060}   &  \bf{267.90}   \\
\hline
\bf{BAIR Push Small}   &     &     &    \\
\hline
SAVP     &  0.502   & 0.090   &  280.32   \\
VRNN   & \bf{0.825} & \bf{0.054}  & \bf{148.51}   \\
SVG-LP   &  0.822  & 0.059  & 158.80 \\
CCOT-GAN &  0.723  & 0.063  &  201.72  \\
KCCOT-GAN &  0.765   & 0.060  &  167.94 \\
\hline
\bf{Moving MMNIST}   &     &     &   \\
\hline
SAVP     &  0.571   & 0.123   &  129.33   \\
VRNN   &  0.770   & 0.116   & \bf{59.14}  \\
SVG-LP   &  0.668   & 0.160  & 101.39  \\
CCOT-GAN     &  0.661   & 0.139   &  74.20  \\
KCCOT-GAN &  \bf{0.788}   & \bf{0.975}   &  60.33 \\
\hline
\hline
\vspace{1pt}
\end{tabular}
\label{tab:compare1}
\end{table}

\section{Discussion} \label{sec:diss}
In the present paper we introduce KCCOT-GAN, the first algorithm for sequence prediction that is based on recently developed modifications of optimal transport specifically tailored for path spaces. For this we build on the results by  \citet{DBLP:conf/nips/XuWMA20}, where COT was first applied for the task of sequential generation. Our experiments show the ability of KCCOT-GAN to not only capture the spatial structure in the frames, but also learn the complex dynamics evolving over time.  

A limitation of the KCCOT-GAN algorithm is the restricted sample variations in comparison to the baseline models that emphasize stochastic components in the model design.  An improvement on KCCOT-GAN could be achieved by encoding more stochasiticity.
Another direction for future work is to explore alternative choices of the kernel function convoluted over the empirical measures as well as a bandwidth parameter that better satisfies the conditions required for the convergence guarantee.
One may also construct a learned kernel in a similar manner as done in MMD-GAN \cite{li2017mmd}, whose parameters are updated along with those in the generator and discriminator.

\newpage

\bibliography{reference}

\newpage

\appendix

\begin{center}
\Large {\textbf{Conditional COT-GAN for Video Prediction with Kernel Smoothing: \\Supplementary material}}
\end{center}

\section{Details on regularized Causal Optimal Transport} \label{appx:cotgan}
\subsection{Sinkhorn algorithm}

The entropy-regularized transport problems \eqref{eq:Wce} is obtained by considering an entropic constraint.  
For transport plans with marginals $\mu$ supported on a finite set $\{x^i\}_i$ and $\nu$ on a finite set $\{y^{j}\}_j$, any $\pi\in\Pi(\mu,\nu)$ is also discrete with support on the set of all possible pairs $\{(x^i,y^{j})\}_{i,j}$. Denoting $\pi_{ij}=\pi(x^i,y^{j})$, the Shannon entropy of $\pi$ is given by
$
\textstyle{H(\pi):= -\sum_{i,j} \pi_{ij}\log(\pi_{ij})}.
$
A transport plan in the discrete case can be considered as a table identified with a joint distribution.  The intuition of imposing such a regularization is to restrict the search of couplings to tables with sufficient smoothness in order to improve efficiency. 

When the measures are discrete, such a regularized optimal transport problem becomes easily solvable by using the Sinkhorn algorithm for a given number of iterations, say $L$, in order to approximate a solution to the Sinkhorn divergence \eqref{Sink}, see \cite{GPC} for detail. Generally speaking, the stronger the regularization is (that is, the bigger the parameter $\eps$ is), the fewer number of iterations $L$ is needed in order to yield a good approximation.

\subsection{Sinkhorn divergence at the level of mini-batches}

To correct the fact that $\mathcal{W}_{c,\eps}(\alpha,\alpha)\neq0$, the Sinkhorn divergence proposed by \citet{GPC} at the mini-batch level is written as
\begin{equation} \label{eq:mini-sink}
    \widehat{\mathcal{W}}_{c,\epsilon} (\widehat{\mu},\widehat{\nu}_\theta) := \mathcal{W}_{c,\eps}(\widehat{\mu},\widehat{\nu}_\theta) - \mathcal{W}_{c,\eps}(\widehat{\mu},\widehat{\mu}) - \mathcal{W}_{c,\eps}(\widehat{\nu}_\theta,\widehat{\nu}_\theta),
\end{equation}
where the empirical measures $\widehat{\mu}$ and $\widehat{\nu}_\theta$ correspond to mini-batch sampled from the dataset and that produced by the model, respectively. 

This is an attempt to correct the bias introduced by the entropic regularization via 
eliminating the differences brought by the variations  in both mini-batches of the real and generated samples.  
However, an experiment in \cite{DBLP:conf/nips/XuWMA20} shows that the above formulation \eqref{eq:mini-sink} failed to reduce the bias and recover the optimizer set up as a known quantity. Therefore, the authors propose the mixed Sinkhorn divergence, 
\begin{align*} \label{eq:mini-mix}
    \widehat{\mathcal{W}}_{c,\epsilon}^{\text{mix}} (\widehat{\mu},\widehat{\mu}', \widehat{\nu}_\theta ,\widehat{\nu}'_\theta) := & \mathcal{W}_{c,\eps}(\widehat{\mu},\widehat{\nu}_\theta) 
    + \mathcal{W}_{c,\eps}(\widehat{\mu}',\widehat{\nu}'_\theta)\\
    & - \mathcal{W}_{c,\eps}(\widehat{\mu},\widehat{\mu}') 
    - \mathcal{W}_{c,\eps}(\widehat{\nu}_\theta,\widehat{\nu}'_\theta),
\end{align*}
where $\widehat{\mu}$ and $\widehat{\mu}'$ correspond to different mini-batches from the dataset, and $\widehat{\nu}$ and $\widehat{\nu}'$ from generated samples. Instead of considering the variations within a batch,  the mixed Sinkhorn divergence reduces the bias by excluding the variations in different mini-batches from the same underlying distribution. 

Alternative mini-batch Sinkhorn divergences are also investigated in \cite{DBLP:conf/nips/XuWMA20},  for example,
\begin{align*}
\widehat{\mathcal{W}}_{c,\epsilon}^{6} (\widehat{\mu}, ,\widehat{\mu}', \widehat{\nu}_\theta,\widehat{\nu}'_\theta) 
& = \mathcal{W}_{c, \eps}(\widehat{\mu},\widehat{\nu}_\theta)
+\mathcal{W}_{c, \eps}(\widehat{\mu}',\widehat{\nu}_\theta) \\
& +\mathcal{W}_{c, \eps}(\widehat{\mu},\widehat{\nu}'_\theta) 
\quad +\mathcal{W}_{c, \eps}(\widehat{\mu}',\widehat{\nu}'_\theta) \\
& -2\mathcal{W}_{c, \eps}(\widehat{\mu}',\widehat{\mu}')
-2\mathcal{W}_{c, \eps}(\widehat{\nu},\widehat{\nu}'_\theta). 
\end{align*}
In sequential generation (without conditioning), the results in \cite{DBLP:conf/nips/XuWMA20} suggest that $\widehat{\mathcal{W}}_{c,\epsilon}^{\text{mix}}$ and $\widehat{\mathcal{W}}_{c,\epsilon}^{6}$ outperform all other formulations of mini-batch Sinkhorn divergence in both the low-dimensional experiments and video generation.   
Although $\widehat{\mathcal{W}}_{c,\epsilon}^{\text{mix}}$ and $\widehat{\mathcal{W}}_{c,\epsilon}^{6}$ produce equally good results, $\widehat{\mathcal{W}}_{c,\epsilon}^{6}$ is 
computationally more expensive because it requires two more terms in the computation.

In the case of sequential prediction, $\widehat{\mathcal{W}}_{c,\epsilon} (\widehat{\mu},\widehat{\nu}_\theta)$ is employed in the KCCCOT-GAN algorithm.  
Recall that $\widehat{\nu}_\theta$ denotes the empirical measure of the concatenated sequences which share the input sequences with the real sequences up to time step $k$. As a result, it is not sensible to account for the variations in two batches from the same distribution that do not coincide before time step $k$ as $\widehat{\mu}$ and $\widehat{\nu}_\theta$ do. 
Hence, we consider $\widehat{\mathcal{W}}_{c,\epsilon} (\widehat{\mu},\widehat{\nu}_\theta)$ a more appropriate objective function for prediction under the setting of KCCOT-GAN. 

\subsection{An equivalent characterization of causality} \label{appx:causal_eq}
The expression \eqref{Pcce} obtained in Section~\ref{section:cCOT-GAN} relies on the following characterization of causality, proved in \cite{backhoff2017causal}: a transport plan $\pi\in\Pi(\mu,\nu)$ is causal if and only if
\begin{equation}\label{causalhM}
\textstyle{\E^{\pi}\left[\sum_{t=1}^{T-1} h_t(y) \Delta_{t+1}M(x)\right] = 0\;\; \text{for all $(h,M)\in\Hcal(\mu)$}}.
\end{equation}
With an abuse of notation we write $h_t(y)$, $M_t(x)$, $\Delta_{t+1} M(x)$ rather than $h_t(y_{1:t})$, $M_t(x_{1:t})$, $\Delta_{t+1} M(x_{1:t+1})$.

\subsection{Details about COT-GAN}
Adopting the mixed Sinkhorn divergence, COT-GAN is trained on the following objective function
\begin{equation} \label{eq:wmix_obj}
    \widehat{\mathcal{W}}_{c,\epsilon}^{\text{mix}, L} (\widehat{\mu},\widehat{\mu}', \widehat{\nu}_\theta, \widehat{\nu}'_\theta) - \lambda p_{\vM_{\varphi_2}} (\widehat{\mu}),
\end{equation}
where $L$ indicates the number of iterations required for approaching a solution to the mixed Sinkhorn divergence. 

To formulate an adversarial training algorithm for implicit generative models,  
COT-GAN approximates the set of functions \eqref{eq:L_set} by truncating the sums at a fixed $J$, 
and parameterizes $\vh_{\varphi_1}:=(h_{\varphi_1}^j)_{j=1}^J$ and 
$\vM_{\varphi_2}:=(M_{\varphi_2}^j)_{j=1}^J$ as two separate 
neural networks, and let $\varphi:=(\varphi_1,\varphi_2)$.  To capture the characteristics of those processes, the choices of network architecture are restricted to those with causal connections only. The mixed Sinkhorn divergence is then calculated with respect to a parameterized cost function
\begin{equation}\label{eq:cot_cost}
    c^\Kcal_{\varphi}(x,y):=c(x,y) + \sum_{j=1}^J\sum_{t=1}^{T-1} h^{j}_{\varphi_1,t}(y) \Delta_{t+1}M^{j}_{\varphi_2}(x),
\end{equation}
where the cost function is chosen to be $c(x,y)=\| x- y \|^2_2$ in COT-GAN. 

While the generator $g_\theta:\mathcal{Z}\to\mathcal{X}$ is incorporated in $\widehat{\nu}_\theta$, the discriminator role in COT-GAN is played by $\vh_{\varphi_1}$ and $\vM_{\varphi_2}$. COT-GAN learns a robust (worst-case) distance between the real data distribution and the generated distribution by maximizing the objective \eqref{eq:wmix_obj} over $\varphi$, and a strong generator to fool the discriminator by minimizing the mixed divergence over $\theta$. 

\section{Experiment details} \label{appx:experiments}
\subsection{Network architectures and training details} \label{appx:architectures}
\begin{table} [t]
\centering
\caption{Encoder and decoder architecture.}
 \begin{tabular}{||c c||}
 \hline
 \hline
  & Encoder Configuration  \\ 
 \hline\hline
 Input & $x_{1:T}$ with shape $T \times 64 \times 64 \times 3$  \\
 \hline
 1 & convLSTM2D(N32, K6, S2, P=SAME), LN \\
 \hline
 2 & convLSTM2D(N64, K6, S2, P=SAME), LN \\
 \hline
 3 & convLSTM2D(N128, K5, S2, P=SAME), LN \\
 \hline
 4 & convLSTM2D(N256, K5, S2, P=SAME), LN \\
 \hline
 5 & output features $e_{1:T}$ with shape $T \times 4 \times 4 \times 256$ \\
 \hline
 \hline
  & Decoder Configuration  \\ 
 \hline
 \hline
 Input &  $z_{k:T-1}$, $e_{k:T-1}$, $x_{k:T-1}$   \\ 
 \hline
 1 & DCONV(N256, K2, S2, P=SAME), LN \\
 \hline
 2 & convLSTM2D(N128, K4, S1, P=SAME), LN\\
 \hline
 3 &  DCONV(N128, K4, S2, P=SAME), LN \\
 \hline
 4 & convLSTM2D(N64, K6, S1, P=SAME), LN\\
 \hline
 5 & DCONV(N64, K6, S2, P=SAME), LN \\
 \hline
 6 & convLSTM2D(N32, K6, S1, P=SAME), LN\\
 \hline
 4 & DCONV(N16, K6, S1, P=SAME), LN\\
 \hline
 5 & convLSTM2D(N8, K8, S1, P=SAME), LN \\
 \hline
 7 & DCONV(N3, K8, S1, P=SAME), Sigmoid \\
 \hline \hline
\end{tabular}
\label{table:g_structure}
\end{table}
All experiments on the three datasets share the same GAN architectures. The generator is split into an encoder and a decoder, supported by convolutional LSTM (convLSTM). The encoder learns both the spatial and temporal features of the input sequences, whereas the decoder predicts the future evolution conditioned on the learned features and a latent variable.  

The features from the last encoding layer has a shape of $4 \times 4 $ (height $\times$ width) per time step.  A latent variable $z$ is sampled from a multivariate standard normal distribution with the same shape as the features (same number of channels too depending on the model size). We then concatenate the features, input sequence, and latent variables over the channel dimension as input for the decoder.  The encoder and decoder structures are detailed in Table \ref{table:g_structure}.
As the discriminator, the process $\vh$ and $\vM$ are parameterized with two separate networks that share the same structure, shown in Table \ref{table:d_structure}. 
In all tables, we use DCONV to represent a de-convolutional (convolutional transpose) layer.  The layers may have N filter size, K kernel size, S strides and P padding option.  We adopt both batch-normalization(BN) and layer-normalization(LN), and the LeakyReLU activation function. All hyperparameter setting are the same for all three datasets except that the filter size is halved for the Moving MNIST dataset.
\begin{table} [!]
\caption{Discriminator architecture.}
\centering
 \begin{tabular}{||c c||}
 \hline \hline
 Discriminator & Configuration  \\ 
 \hline\hline
 Input & 64x64x3 \\
 \hline
 0 & CONV(N32, K5, S2, P=SAME), BN \\
 \hline
 1 & CONV(N64, K5, S2, P=SAME), BN \\
 \hline
 2 & CONV(N128, K5, S2, P=SAME), BN \\
 \hline
 3 & reshape 3D array for LSTM \\ 
 \hline
 4 & LSTM(state size = 128), LN \\ 
 \hline
 5 & LSTM(state size = 64), LN  \\ 
 \hline
 6 & LSTM(state size = 32), LN  \\ 
 \hline  \hline
\end{tabular}
\label{table:d_structure}
\end{table}

During training, we apply exponential decay to the learning rate by $\eta_t = \eta_0 r^{s/c}$ where $\eta_0$ is the initial learning rate, $r$ is decay rate, $s$ is the current number of training steps and $c$ is the decaying frequency.  The bandwidth parameter $h$ are also annealed from $1.5$ to $0.1$ in a similar manner. 
In all experiments, the initial learning rate is $0.0005$, decay rate $0.985$, decaying frequency $10000$, and batch size $m=8$. The settings of hyper-parameters in the Sinkhorn algorithm are also shared across the three datasets with $\lambda=1.0$, $\eps = 0.8$ and the Sinkhorn iterations $L=100$.  We train KCCOT-GAN and CCOT-GAN on a single NVIDIA GTX 1080 Ti GPU. Each iteration takes roughly 3.5 seconds. Each experiment is run for around 100000 iterations.

\subsection{Results on Moving MNIST}
\begin{figure*}[t!]
    \centering
    \includegraphics[width=1.0\textwidth]{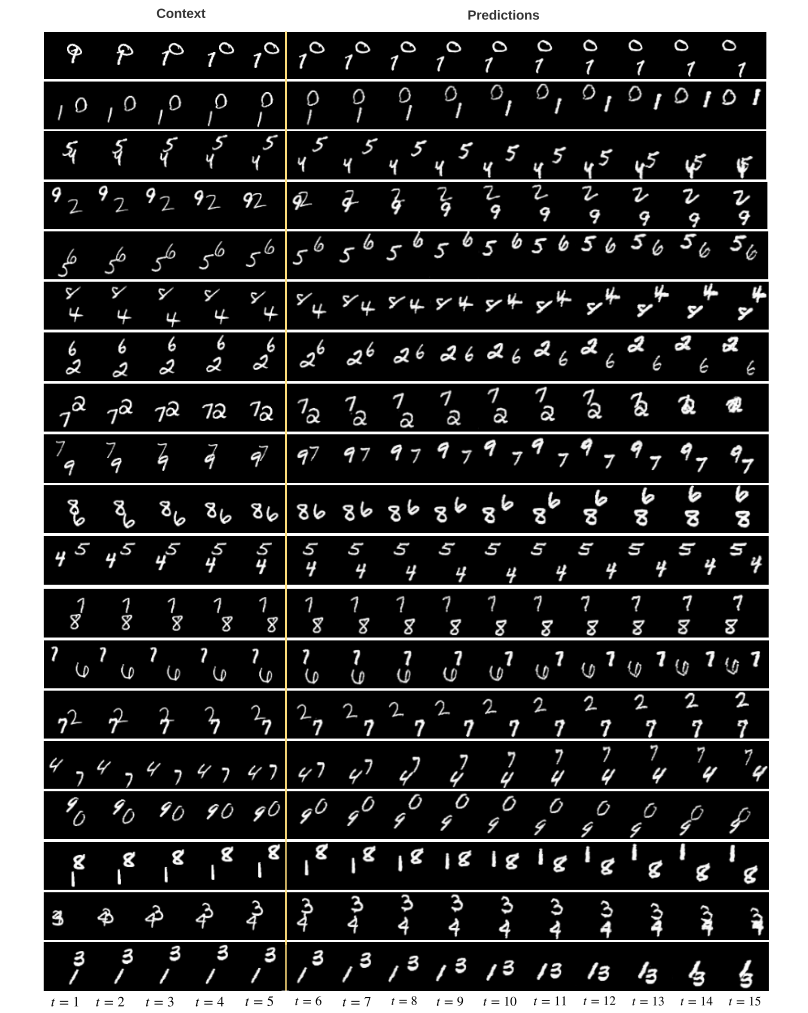}
    \caption{Moving MNIST results on test set. The first 5 frames are context sequence and last 10 frames are predictions from KCCOT-GAN, separated by the yellow vertical line.  }
    \label{fig:mmnist}
\end{figure*}

Predictions from KCCOT-GAN conditioned on the first 5 context frames from the test set of the Moving MNIST dataset are presented in Figure \ref{fig:mmnist}. 
\end{document}